\documentclass[runningheads]{llncs}
\usepackage{graphicx}
\usepackage{multirow}
\usepackage{amsmath}
\usepackage{listings}
\usepackage{changepage}
\usepackage{xcolor}

\begin{document}
\title{Few-shot medical image classification with simple shape and texture text descriptors using vision-language models}

\titlerunning{Few-shot classification with simple text descriptors}
%


\author{Michal Byra\inst{1,2,}\thanks{Corresponding author.}, 
Muhammad Febrian Rachmadi\inst{1,3} \and
Henrik Skibbe\inst{1}}
\authorrunning{M. Byra et al.}
%


\institute{RIKEN Center for Brain Science, Wako, Japan \and
Institute of Fundamental Technological Research,  \\ Polish Academy of 
Sciences, Warsaw, Poland \and
Faculty of Computer Science, Universitas Indonesia, Depok, Indonesia \\ \email{michal.byra@riken.jp} }

%
\maketitle              
\begin{abstract}

In this work, we investigate the usefulness of vision-language models (VLMs) and large language models for binary few-shot classification of medical images. We utilize the GPT-4 model to generate text descriptors that encapsulate the shape and texture characteristics of objects in medical images. Subsequently, these GPT-4 generated descriptors, alongside VLMs pre-trained on natural images, are employed to classify chest X-rays and breast ultrasound images. Our results indicate that few-shot classification of medical images using VLMs and GPT-4 generated descriptors is a viable approach. However, accurate classification requires to exclude certain descriptors from the calculations of the classification scores. Moreover, we assess the ability of VLMs to evaluate shape features in breast mass ultrasound images. We further investigate the degree of variability among the sets of text descriptors produced by GPT-4. Our work provides several important insights about the application of VLMs for medical image analysis. 

\keywords{medical image classification  \and vision-language models \and large language models \and few-shot learning}
\end{abstract}

\section{Introduction}

Vision-language models (VLM) and large language models are gaining momentum in machine learning. VLMs trained on paired image-text data have been successfully used for 0-shot classification, image-to-text matching and object detection, among many other tasks \cite{zhang2023vision}. In medical image analysis, VLMs have been mainly applied for chest X-ray images analysis due to the public availability of large datasets of radiology reports paired with imaging data, such as the MIMIC-CXR \cite{johnson2019mimic}. For example, Keicher et al. utilized VLMs pre-trained on MIMIC-CXR to automate the reporting and assessment of pathologies in chest X-ray images. Boecking et al. developed BioVil, a large VLM pre-trained on chest X-rays and radiology reports \cite{boecking2022making}. Authors presented that the developed model can be used for various downstream tasks, including  0-shot classification and image-text retrieval. In the standard setting, the 0-shot image classification with VLMs, such as the CLIP model, is performed by calculating similarity scores between the input image and text descriptors designed for different classification categories, such as "a photo of a cat"  \cite{Radford2021LearningTV}. Recently, Menon and Vondrick presented that 0-shot classification can be performed based on text descriptors characterizing image features associated with the class categories. Authors used GPT-3 to automatically generate the descriptors. For example, classification of an airliner in images can be performed with VLMs using text descriptors related to the properties of the airliner, such as ``large, metal aircraft". This approach is explainable by design, as the classification decision can be justified by the presence of particular image features. Building upon this approach and the BioVil VLM, Pellegrini et al. proposed Xplainer, an explainable 0-shot classification method for chest X-rays \cite{pellegrini2023xplainer}. Authors prompted ChatGPT to provide radiology report like text descriptors for the assessment of various chest pathologies. Qin et al. utilized GLIP model for object detection based on medical prompts \cite{Qin2022MedicalIU}. 

In radiology, the classification is sometimes conducted based on the presence of simple image features. For example, to differentiate malignant and benign breast masses in ultrasound (US), radiologists assess the texture and shape characteristics of the lesions \cite{flores2015improving}. Standard reporting includes the evaluation of the roundness, mass contour variability and mass echogenicity. In this work, building upon the study of Menon and Vondrick, we investigate if VLMs and large language models can be used for the binary few-shot classification of medical images. 
In order to address the particular task of medical image classification, we prompt GPT-4 to generate simple plain text descriptors related to the shape and texture of objects in chest X-rays and breast US images. These descriptors, in conjunction with VLMs pre-trained on natural images, are then utilized for image classification. For example, to differentiate between malignant and benign breast masses, we employ descriptors such as ``round shape" or ``variable texture". Our results demonstrate the feasibility of few-shot classification of medical images using VLMs and GPT-4-generated descriptors. This approach distinguishes itself from previous works on zero-shot medical image classification by eliminating the need for training the VLM on specific datasets of paired clinical reports and medical images. In addition, we investigate the ability of the VLMs to accurately assess shape features in medical images and evaluate the variability in the sets of text descriptors generated by GPT-4.

\section{Methods}

\subsection{Generating text descriptors}

\begin{figure}[t!]
	\begin{center}
		\includegraphics[width=0.75\linewidth]{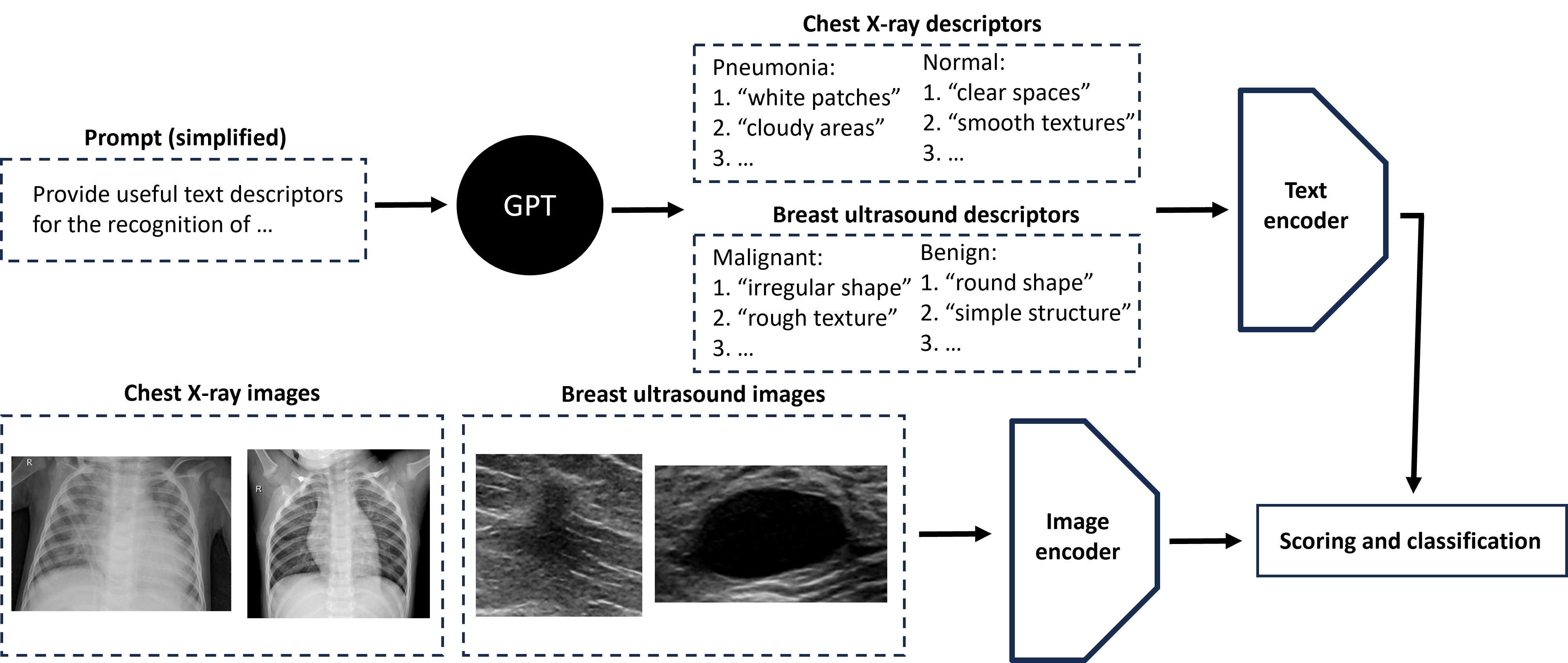}
	\end{center}
	\caption{Scheme presenting the proposed approach to few-shot medical image classification with vision-language models. GPT-4 was used to generate simple text descriptors related to the shape and texture of medical images. }
	\label{fig:f1}
\end{figure}

Figure \ref{fig:f1} presents our approach to the few-shot classification of medical images. First, we used GPT-4 to generate suitable text descriptors for the particular classification task. GPT-4 was prompted to generate 20 simple descriptors related to shape and texture features. These descriptors should enable a VLM, pre-trained on natural images, to effectively handle medical images. A separate set of 20 text descriptors was generated for each class category. See Appendix A for the exemplary GPT-4 prompts and the generated text descriptors.

\subsection{Classification}

The generated text descriptors and the images were inputted to the VLM to determine the text-image similarity. 0-shot classification was performed based on the class score function, which has the following form \cite{menon2022visual}: 

\begin{equation}
    s(c,x) = \frac{1}{|D(c)|} \sum_{d \in D(c)} \phi (d,x)
\end{equation}

\noindent where $D(c)$ is the set of descriptors corresponding to class $c$ and $\phi (d,x)$ stands for the VLM output (dot product based similarity score) determined for the text descriptor $d$ and image $x$. The class score function should be high for the descriptors that accurately pertain to the input image. In our study, to perform the binary classification, we calculated the following classification score function: 

\begin{equation}
    p(x) = s(c=1,x) - s(c=-1,x),
\end{equation}

\noindent where we followed the convention that labels for the positive and negative classes are coded with 1 and -1, respectively. Input image is categorized as belonging to the positive class when $p(x)$ exceeds a specific threshold $b$, $p(x)>b$. This classification cut-off can be set based on training data to provide required sensitivity and specificity. For the 0-shot classification, the cut-off $b$ is simply set to 0 \cite{menon2022visual}.  

\subsection{$n$-shot descriptor selection}

GPT-4 may generate descriptors that are not suitable for medical image analysis. Moreover, specific text descriptors may not work well with the VLMs pre-trained on natural images. To address this problem, we utilized an $n$-shot descriptor selection method. Given only several pairs ($n>0$) of images corresponding to positive and negative classes, our goal was to exclude the worse performing descriptors from the calculations of the sum in eq. 1. To achieve this, we utilized the following descriptor score function:   

\begin{equation}
    r(d_c) = \frac{1}{|X|} \sum_{i=1}^{|X|} c_i \phi (d_c,x_i),
\end{equation}

\noindent where $c \in \{-1,1\}$, and $X$ is the training set of image pairs. The subscript $c$ in $d_c$ indicates that descriptor $d_c$ was generated for class $c$. In the ideal case, the output values of the VLM should be larger for the images corresponding to the target class than for the other classes. Hence for well designed descriptors, we expect that the scoring function $r(d_c)$ is positive. Given a small training set $X$, we exclude from the sum, eq. 1, the descriptors for which the scoring function $r(d_c)$ is negative. Next, using the selected descriptors, we modify eq.1 and formulate the weighted category score as follows: 

\begin{equation}
    s'(c,x) = \frac{1}{\sum_{d \in D'(c)} r(d)} \sum_{d \in D'(c)} r(d) \phi (d,x), 
\end{equation}

\noindent which corresponds to the arithmetic mean weighted based on the scoring function $r(d)$. $D'(c)$ is the pruned set of descriptors after the removal of the descriptors with $r(d)<0$. Moreover, the weighted category score is then used to calculate the classification score in eq. 2.

\subsection{Datasets and implementation}

Classification performance was assessed using accuracy and the area under the receiver operating characteristic curve (AUC) calculated using classification scores. Cut-off $b$ for the calculation of the accuracy in the few-shot setting, eq. 2, was selected based on the AUC to optimize the UL index. We used the following datasets for the experiments: 

\subsubsection{Chest X-ray.} We utilized a public chest X-ray dataset, consisting of 5856 cases~\cite{kermany2018identifying}. 4273 images corresponded to the pneumonia and 1583 to normal X-rays. For the evaluation, we used the training/test split provided by the authors, with the test set including 390 pneumonia images and 234 normal chest X-rays. 

\subsubsection{Breast ultrasound.} We used the UDIAT dataset, consisting of 159 US images (4 duplicated US images were removed) corresponding to 107 benign and 52 malignant breast masses \cite{yap2017automated}. Each US image had a breast mass area segmentation mask outlined by an expert. The dataset was divided into training/test sets with a 104/55 split, with the ratio of the malignant and benign masses maintained for both sets. In addition, the US images were cropped based on the segmentation masks with a margin of 20 pixels.    

Calculations were performed in PyTorch \cite{ilharco_gabriel_2021_5143773,NEURIPS2019_9015,Radford2021LearningTV}. We utilized the openAI's CLIP ViT-bigG/14 VLM pre-trained on the LAION-2B dataset~\cite{schuhmann2022laionb}. Text descriptors and  images were pre-processed with the routines designed for the model. Code is available at github.com/BrainImageAnalysis/FSC-CLIP-GPT.   

\section{Experiments}

\subsection{Shape assessment}

\begin{figure}[]
	\begin{center}
		\includegraphics[width=0.7\linewidth]{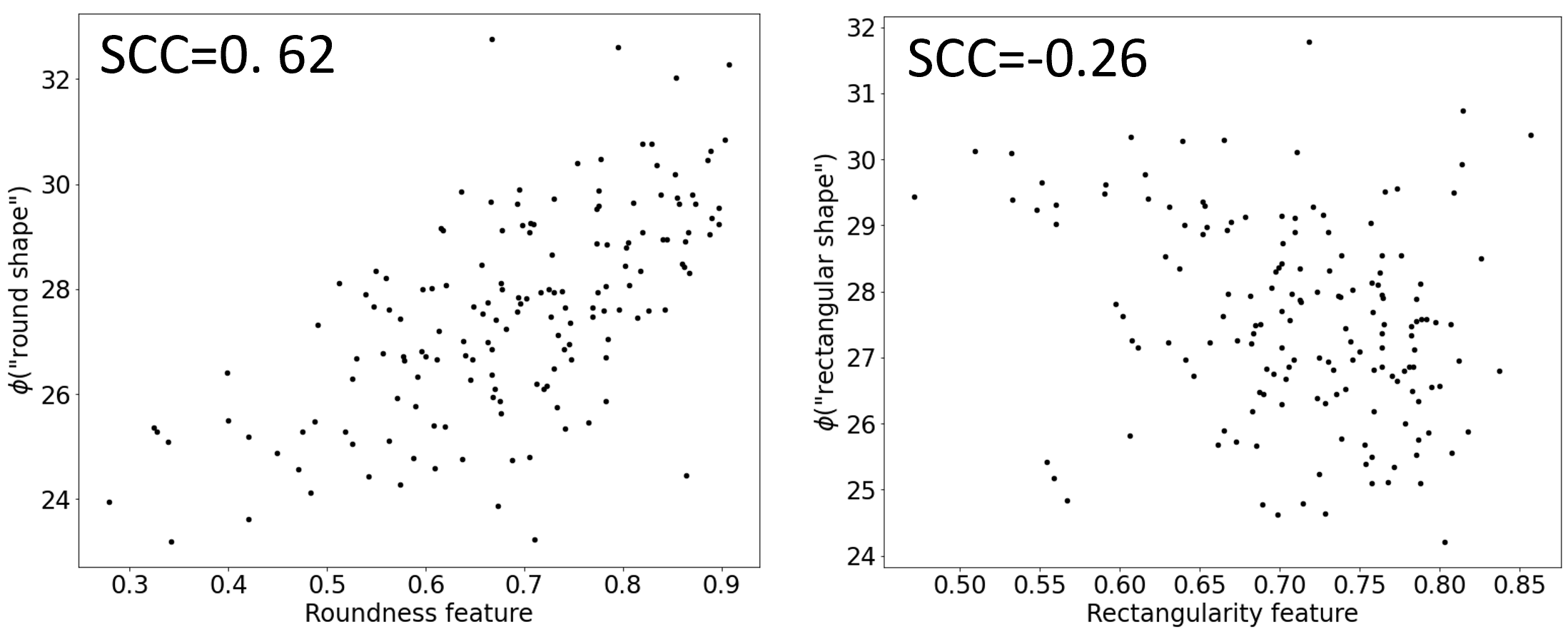}
	\end{center}
	\caption{The relationship between the shape features calculated using segmentation masks and the outputs of the vision-language model for the text descriptors related to each shape parameter.}
	\label{fig:f2}
\end{figure}

We examined how accurately the VLM can assess the shape of the breast masses. For this task, we used the segmentation masks to calculate the roundness and rectangularity features. Next, the computed shape features were compared with the outputs of the VLM obtained for the descriptors ``round shape" and ``rectangular shape", respectively. Fig. \ref{fig:f2} shows good correspondence obtained for the roundness parameter, with Spearman's rank correlation coefficient (SCC) of 0.62. However, the model did not provide good results with respect to the rectangularity feature, SCC of -0.26. This result suggest that the capabilities of the VLMs to assess certain image features may be limited in practice. Additional results can be found in Appendix B, where we examined different roundness related text descriptors and concluded that simple descriptors (e.g. made of single words) are the most suitable for the shape assessment with VLMs.   

\subsection{Classification}

\begin{figure}[]
	\begin{center}
		\includegraphics[width=0.6\linewidth]{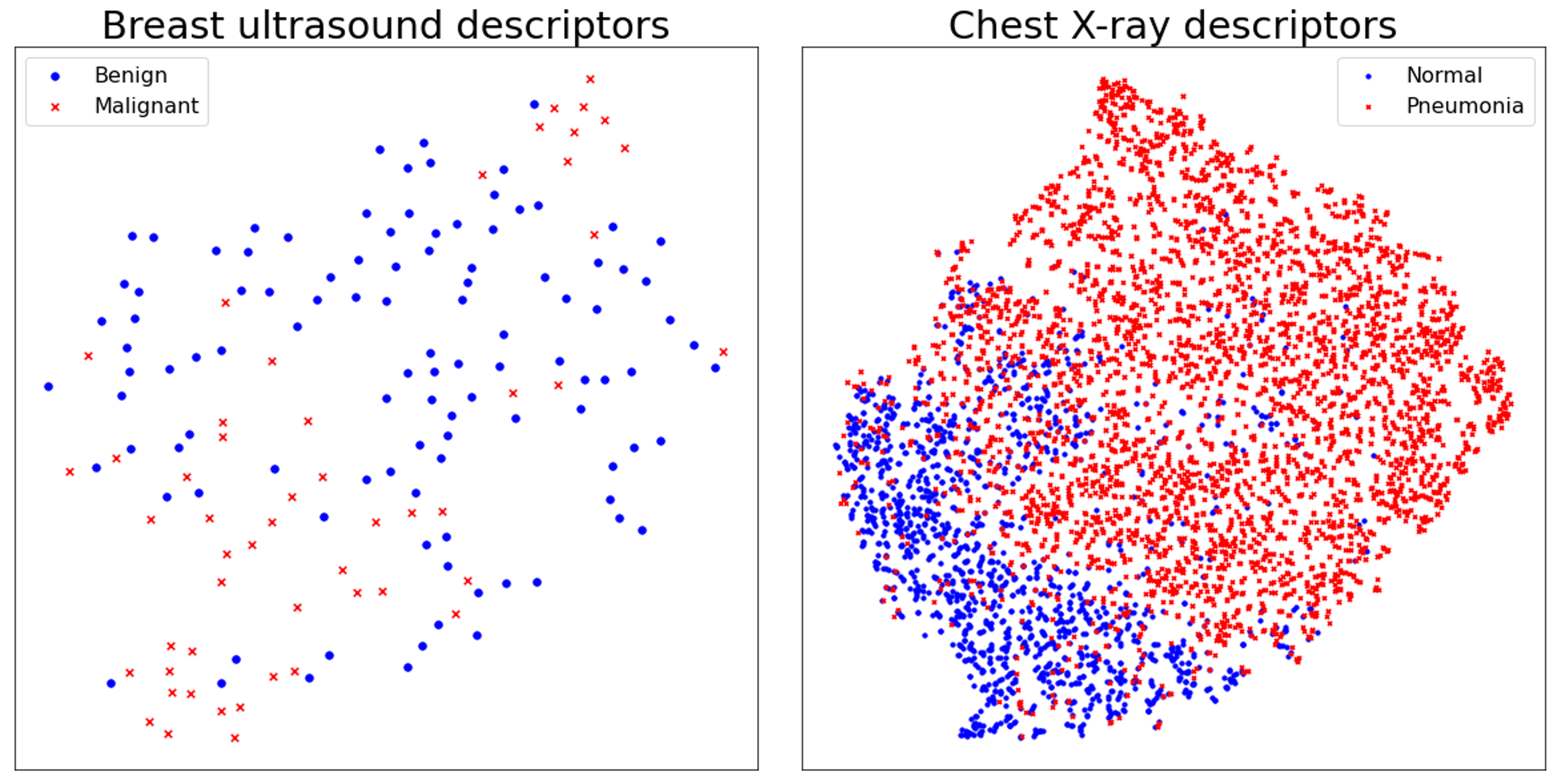}
	\end{center}
	\caption{t-SNE 2D embedding graphs presenting the separability of the classes in breast ultrasound images and chest X-rays. Each embedding was computed based on outputs of the vision-language model using GPT-4 generated text descriptors.}
	\label{fig:f3}
\end{figure}

For each dataset, we joined the VLM outputs corresponding to 20 descriptors for each class to form a 40 feature vector. Next, we applied the t-SNE algorithm to visualise the class separation in the 2D embedding space. Figure \ref{fig:f3} presents the results obtained for the breast US images and chest X-rays, which confirm that the VLMs has the capabilities to differentiate pathologies in medical images based on simple GPT-4 generated text descriptors.   

\begin{figure}[b!]
	\begin{center}
		\includegraphics[width=0.7\linewidth]{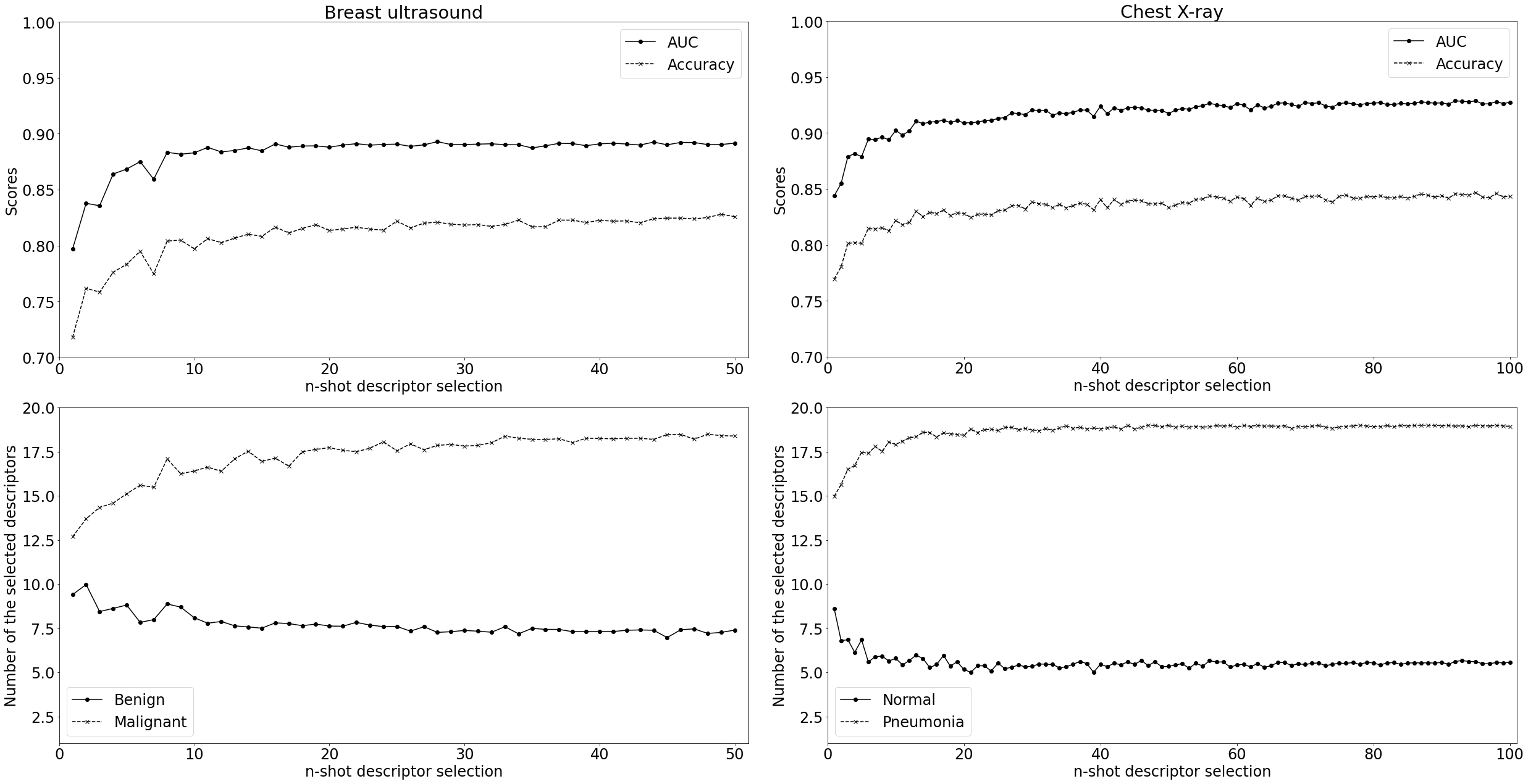}
	\end{center}
	\caption{Classification performance obtained with the proposed $n$-shot ($n>0$) descriptor selection method, with $n$ indicating the number of the image pairs used for the selection.}
	\label{fig:f4}
\end{figure}

Table \ref{tab:t1} presents the classification performance obtained on the test set. Using GPT-4 generated descriptors and the VLM, we obtained good 0-shot classification performance for the chest X-rays, with accuracy and AUC of 0.79 and 0.88, respectively. For the breast masses the accuracy was low and equal to 0.33 while the AUC value was high and equal to 0.89, which suggests that the default 0-shot classification cut-off of 0 was not suitable for the recognition of malignant breast masses. Utilization of the proposed descriptor selection method and cut-off adjustments addressed this problem and resulted in better performance. Figure~\ref{fig:f4} presents the $n$-shot classification performance obtained for different values of $n$ ($n>0$). Each point of the curve corresponds to the average values of the metrics calculated over 100 runs based on random sampling from the training set. For the chest X-rays, we randomly sampled image pairs without replacement. However, for the breast US data we sampled with replacement due to the small volume of the training set. 

Figure~\ref{fig:f4} also illustrates the average number of the selected descriptors with the proposed $n$-shot method. For example, the optimal text descriptor set corresponding to the pneumonia class included around 6 descriptors. Compared to the 0-shot classification, we obtained slightly lower performance for the 1-shot selection method with respect to the AUC value. Presumably, the descriptor selection based on a single image pair was too random, resulting in accidental removal of the better performing descriptors. However, even with a single image pair it was feasible to fix the cut-off issue for the breast mass classification and increase the accuracy to around 0.72. In general, classification performance increased with the number of cases used for the $n$-shot descriptor selection. For the 20-shot classification, the accuracy metrics increased to around 0.81 and 0.83 for the breast US images and chest X-rays, respectively.  

\begin{table}[t!]
\small
\centering
\caption{Classification performance obtained for the chest X-rays and breast ultrasound images.}
\label{tab:t1}

\begin{tabular}{|l|c|c|c|}
\hline

                Dataset  & n-shot & Accuracy $\uparrow$ & AUC $\uparrow$  \\ \hline
\multirow{3}{*}{Chest X-rays} & 0 & 0.79 & 0.88 \\
                  & 1 & 0.72 & 0.80 \\
                  & 10 & 0.80 & 0.88  \\
                  & 20 & 0.81 & 0.89 \\  \hline
\multirow{4}{*}{Breast US} & 0 & 0.33 & 0.89 \\
                  & 1  & 0.78 & 0.85 \\
                  & 10 & 0.82 & 0.90 \\
                  & 20 & 0.83 & 0.91 \\ \hline
\end{tabular}

\end{table}
 
We also investigated the variability of the text descriptor generation method based on GPT-4. The generation process was repeated 50 times and we removed the requirement to provide 20 descriptors for each class. The 0-shot classification performance was evaluated with the entire datasets. The average AUC values, Table \ref{tab:t2}, for the breast US data and chest X-rays were equal to around 0.72 and 0.76, respectively. An ensemble (average of the classification scores, eq. 2) over 50 runs resulted in AUC values of around 0.81 for both datasets. The agreement with respect to the classification scores determined for the 50 sets was moderate, with the interclass correlations (ICC) equal to 0.55 and 0.68 for the breast and chest datasets, respectively. The average number of the descriptors outputted by GPT-4 was equal to 9.8, 9.6, 11, 10.9 for the malignant masses, benign lesions, pneumonia and normal chest X-rays, respectively. These 0-shot classification results suggest that the descriptor generation process was moderately stable, as lack of the better performing descriptors or not adequate text phrasing could potentially decrease the performance. Our study suggests that for the better performance it may be beneficial to generate a larger set descriptors and remove the redundant and under-performing ones based on the training set, as it is unclear which texture image features can be successfully extracted with the VLM in advance. Additional results listing the most frequently outputted text descriptors by GPT-4 can be found in Appendix~C.

\begin{table}[t!]
\small
\centering
\caption{Variability in 0-shot classification performance determined based on 50 sets of text descriptors generated using GPT-4.}
\label{tab:t2}

\begin{tabular}{|c|c|c|}
\hline

Dataset  & AUC $\uparrow$ (mean, min, max, ensemble) & ICC $\uparrow$      \\ \hline

Chest X-rays &  (0.76, 0.50, 0.94, 0.81) & 0.68 \\ \hline

Breast ultrasound &  (0.72, 0.56, 0.84, 0.81) & 0.55   \\ \hline

\end{tabular}

\end{table}

\subsection{Limitations}

Our work has several limitations. First, we used a model developed using natural images, but a VLM fine-tuned with clinical reports and medical images corresponding to different modalities would probably better serve for the investigated classification problems. Second, we used a generic approach to generate text descriptors based on GPT-4. The descriptors were not curated by domain experts. However, our study provides some insights on how to engineer text descriptors to obtain good results with the VLMs. For our few-shot learning methods, we obtained worse  performance compared to previous studies utilizing networks trained in supervised manner, reporting, for example, AUC values of 0.97 for the pneumonia classification \cite{kermany2018identifying}. It remains to investigate whether additional text descriptors can be incorporated to further improve the performance. 

\section{Conclusion}

Establishing the feasibility of using vision-language models for few-shot classification of medical images is a critical step toward broader application of foundation models in medical image analysis. Our research unveiled the effectiveness of employing GPT-4-generated descriptors, associated with features in chest X-rays and breast mass ultrasound images, for this task.

The necessity of careful descriptor selection was underscored by our findings, particularly as the exclusion of certain descriptors was found to be vital for good classification performance. Our evaluations demonstrated the potential of the vision-language models, with the noteworthy accuracy of 0.81 and 0.83 for the X-rays and ultrasound images, respectively, providing encouraging evidence of the applicability of our approach. Nonetheless, the variability observed in the descriptor generation process signifies room for improvement.

\noindent \textbf{Acknowledgement.} The authors do not have any conflicts of interest. This work was supported by the program for Brain Mapping by Integrated Neurotechnologies for Disease Studies (Brain/MINDS) from the Japan Agency for Medical Research and Development AMED (JP15dm0207001) and the Japan Society for the Promotion of Science (JSPS, Fellowship PE21032).


%
\bibliographystyle{splncs04}
\bibliography{mybibliography}

\newpage
\appendix

\section{GPT-4 prompts}

The following prompts were inputted to GPT-4 (chat.openai.com) to obtain the text descriptors for the chest X-rays and breast mass US images. 

\begin{adjustwidth}{0.5cm}{0.5cm}

\textbf{Prompt for chest X-rays}: I am going to use the CLIP vision-language model to detect pneumonia (vs normal) in chest X-ray images. Please, generate 20 text descriptors (text tokens) for the pneumonia differentiation with CLIP. Descriptors should correlate with visual features present in pneumonia images. Next, generate 20 text descriptors for the normal images. The descriptors for the pneumonia should differ from the descriptors for normal X-rays. Each descriptor should consist of simple plain words and refer to shape or texture of the X-ray image. Avoid using words such as ``lung" as the CLIP models were not trained with medical data. 

\noindent \textbf{GPT-4}: Sure, I can provide some plain language descriptors that might be applicable to differentiating pneumonia from normal chest X-ray images... 
\\
\\
\noindent Pneumonia descriptors = ``hazy shadowing", ``opaque areas", ``increased whiteness", ``darker spots", ``blurred contours", ``patchy white areas", ``irregular dense patterns", ``unsmooth texture", ``diffuse bright spots",     ``random white patches", ``cloudy formations", ``inconsistent shadowing", ``whitened irregularities", ``rough textured zones", ``dense irregular patches", ``large dark spots", ``wide white areas", ``vague shadowy shapes", ``dense white clusters", ``irregularly scattered whiteness".
\\
\\
Normal descriptors = 
    ``clear spaces",
    ``consistent gray tone",
    ``smooth textures",
    ``regular patterns",
    ``balanced light and dark areas",
    ``uniform gray shading",
    ``well-defined contours",
    ``few dark spots",
    ``regular shadowing",
    ``lack of white patches",
    ``even grayscale distribution",
    ``smooth gray surfaces",
    ``few bright areas",
    ``symmetrical shading",
    ``uniform transparency",
    ``absence of large dark spots",
    ``evenly distributed grays",
    ``absence of dense white clusters",
    ``uninterrupted gray patterns",
    ``lack of dense patches".
\\
\\
\noindent \textbf{Prompt for breast US images}: I am going to use the CLIP vision-language model to differentiate malignant and benign breast ultrasound images. Please, generate 20 text descriptors for the malignant mass differentiation with CLIP. Descriptors should correlate with visual features present in breast ultrasound images.  Next, generate text descriptors for the benign masses. The descriptors for the malignant mass images should differ from the descriptors for the benign mass images. The descriptors should not be correlated. Each descriptor should consist of simple plain words and refer to shape or texture of the masses in ultrasound images. Avoid using words such as ``mass" or ``cancer" as the CLIP models were not trained with medical data.

\noindent \textbf{GPT-4}: Absolutely, here are your lists of text descriptors... 
\\
\\
\noindent Malignant descriptors = 
   ``irregular outline", 
    ``uneven texture",
    ``sharp edges", 
    ``jagged boundary", 
    ``spiculated periphery",
    ``variegated appearance",
    ``darker shadow",
    ``heterogeneous surface",
    ``oblong form",
    ``dark areas within",
    ``angular shape",
    ``patchy pattern",
    ``lobulated border",
    ``diffuse edges",
    ``taller than wide",
    ``multiple dark spots",
    ``random bright echoes",
    ``hazy border",
    ``mixed echoic pattern",
    ``thick outer line".
    \\
\\
Benign descriptors = 
   ``round shape", 
    ``smooth texture",
    ``circular form", 
    ``consistent pattern", 
    ``soft edges",
    ``uniform appearance",
    ``bright center",
    ``homogeneous surface",
    ``symmetric shape",
    ``uniformly bright echoes",
    ``oval structure",
    ``light shadow",
    ``regular boundary",
    ``spherical configuration",
    ``wider than tall",
    ``single dark spot",
    ``solid light pattern",
    ``clear border",
    ``few echoic areas",
    ``thin outer line".

\end{adjustwidth}

\section{Shape features and text descriptors}

In this study, the roundness feature was calculated based on the manual breast mass segmentations with the following formula:

\begin{equation}
    Roundness = \frac{4 \pi A}{P^2},
\end{equation}

\noindent where $A$ and $P$ indicate the area and the perimeter of the segmentation mask, respectively. Rectangularity feature was computed with the following equation: 

\begin{equation}
    Rectangularity = \frac{A}{A_{bb}}, 
\end{equation}

\noindent where $A$ stands for the mask area and $A_{bb}$ is the area of the bounding box rectangle including the mass. 

\begin{figure}[t!]
	\begin{center}
		\includegraphics[width=1\linewidth]{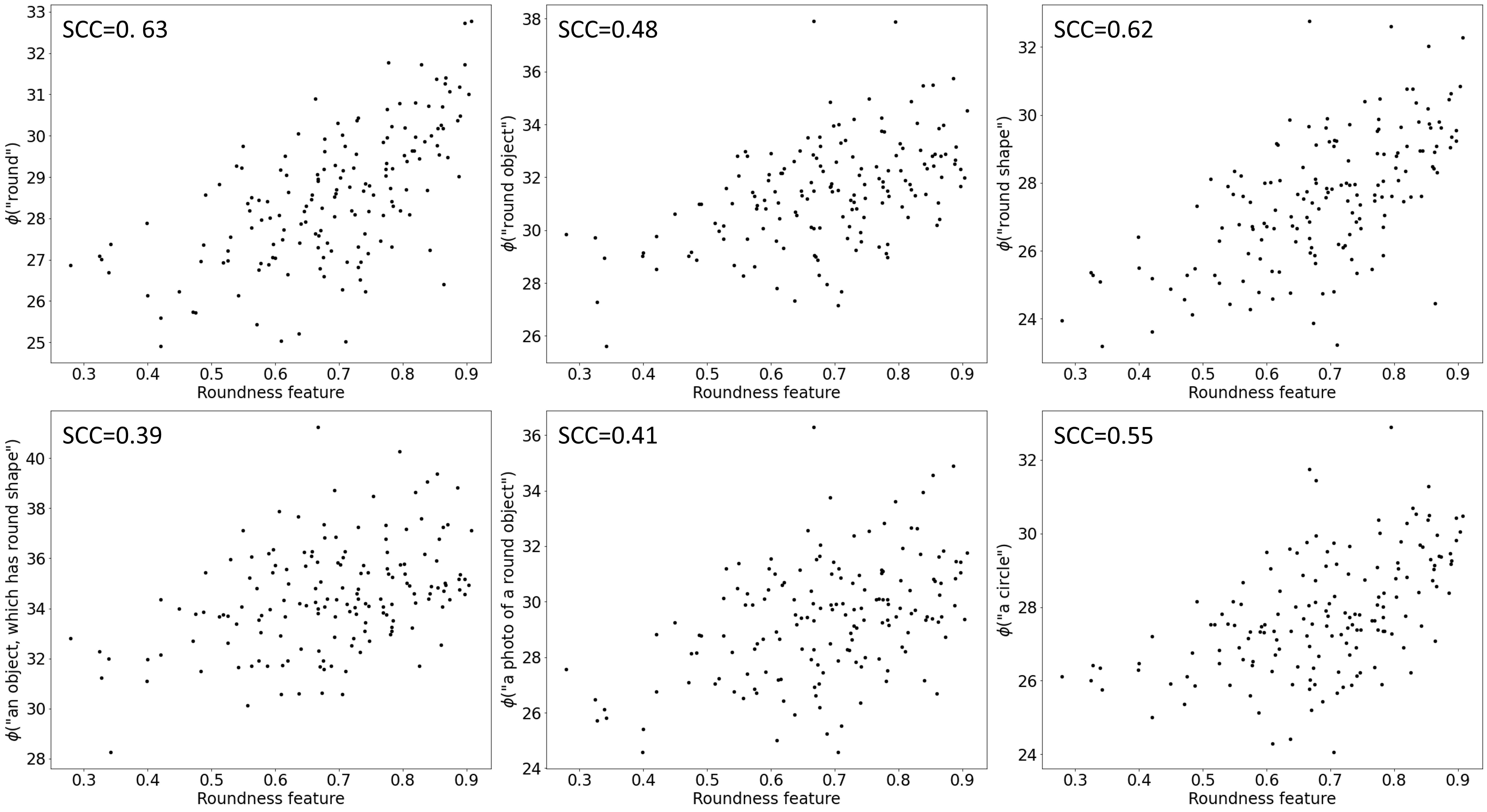}
	\end{center}
	\caption{The relationships between the outputs of the VLM model for different roundness related text descriptors vs the roundness feature calculated based on breast mass segmentation masks. SCC stands for the Spearman's correlation coefficient.}
	\label{fig:f5}
\end{figure}

Additionally, we investigated how the phrasing of the roundness related text descriptor affects the relationship between the VLM output and the roundness feature calculated using segmentation mask. For this, we handcrafted the following six descriptors: ``round", ``round object", ``round shape", ``an object, which has round shape", ``a photo of a round object" and ``a circle". Results presented in Fig. \ref{fig:f5} shows that the better correlation coefficients were obtained for the most simple descriptors, with the ``round" achieving the highest SCC of 0.63. The text descriptor crafting method suggested by Menon and Vondrick, which utilized full sentences, such as ``a photo of a round object", resulted in lower correlation coefficients compared to the basic plain descriptors \cite{menon2022visual}.     

\newpage 

\section{GPT-4 answer variability}

We inputted the prompts described in Appendix A to GPT-4 50 times to assess the variability in the generated text descriptors. For this experiment, we omitted the requirement to provide 20 text descriptors. The most frequently outputted five text descriptor for each class are listed in Table \ref{tab:t3}. Notice that in some cases GPT-4 generated text descriptors related to the same image feature, such as ``irregular shape” and ``irregularly shaped object”. Moreover, for the differentiation between the two classes, GPT-4 provided pairs of opposing descriptors, for example ``cloudy texture" for pneumonia and ``clear texture" for the normal chest X-rays.   

\begin{table}[t!]
\small
\centering
\caption{The most frequently outputted text descriptors by GPT-4.}
\label{tab:t3}
\begin{tabular}{|l|l|c|} \hline
           Class       & Descriptor & Occurrence (max=50) \\ \hline
\multirow{5}{*}{Pneumonia X-rays} & 1. ``cloudy texture" & 21 \\
                  & 2. ``dense spots" & 9 \\
                  & 3. ``diffuse shadows" & 8 \\
                  & 4. ``uneven brightness" & 8 \\
                  & 5. ``blurred boundaries" & 8 \\ \hline
\multirow{5}{*}{Normal X-rays} & 1. ``clear texture" & 15  \\
                  & 2. ``uniform brightness & 12 \\
                  & 3. ``uniform texture" & 10 \\
                  & 4. ``clear image" & 9 \\
                  & 5. ``regular shapes" & 9 \\ \hline
\multirow{5}{*}{Malignant masses} & 1. ``irregular shape" & 16 \\
                  & 2. ``irregularly shaped object" & 16 \\
                  & 3. ``heterogeneous texture" & 8 \\
                  & 4. ``uneven edges" & 7 \\
                  & 5. ``heterogeneous appearance" & 7 \\ \hline
\multirow{5}{*}{Benign masses} & 1. ``round shape" & 12 \\
                  & 2. ``smooth texture" &  9 \\
                  & 3. ``uniform texture" &  8 \\
                  & 4. ``well-defined edges" &  7 \\
                  & 5. ``homogeneous appearance" & 6 \\ \hline
\end{tabular}
\end{table}

\end{document}